\documentclass{article}

    \PassOptionsToPackage{numbers, compress}{natbib}

    \usepackage[preprint]{neurips_2019}

\usepackage[utf8]{inputenc} 
\usepackage[T1]{fontenc}    
\usepackage{hyperref}       
\usepackage{url}            
\usepackage{booktabs}       
\usepackage{amsfonts}       
\usepackage{nicefrac}       
\usepackage{microtype}      
\usepackage{graphicx}
\usepackage{grffile}
\usepackage{float}
\usepackage{caption}
\usepackage{wrapfig}
\usepackage{amsmath}
\usepackage{fixltx2e}
\setcitestyle{square}
\usepackage{tabularx}
\usepackage{multirow}

\usepackage{booktabs,collcell,eqparbox,makecell}
\usepackage{multirow}

 \title{Image Augmentation for Satellite Images}

\author{
 \textbf{Oluwadara Adedeji}\\
  Carnegie Mellon University\\
  Kigali BP 6150, Rwanda\\
  \texttt{oadedeji@andrew.cmu.edu}\\
  \and 
  \textbf{Peter Owoade}\\
   Carnegie Mellon University\\
   Kigali BP 6150, Rwanda\\
  \texttt{powoade@andrew.cmu.edu}\\
  \and 
  \textbf{Opeyemi Ajayi}\\
  Carnegie Mellon University\\
   Kigali BP 6150, Rwanda\\
  \texttt{oajayi@andrew.cmu.edu}\\
  \and 
  \textbf{Olayiwola Arowolo}\\
  Carnegie Mellon University\\
   Kigali BP 6150, Rwanda\\
  \texttt{oarowolo@andrew.cmu.edu}
}

\begin{document}

\maketitle
\begin{abstract}
  This study proposes the use of generative models (GANs) for augmenting the EuroSAT dataset for the Land Use and Land Cover (LULC) Classification task. We used DCGAN and WGAN-GP to generate images for each class in the dataset. We then explored the effect of augmenting the original dataset by about 10\% in each case on model performance. The choice of GAN architecture seems to have no apparent effect on the model performance. However, a combination of geometric augmentation and GAN-generated images improved baseline results. Our study shows that GANs augmentation can improve the generalizability of deep classification models on satellite images.
\end{abstract}

\section{Introduction}

Data augmentation is a popular step in training deep learning models. Data augmentation becomes necessary due to insufficient training data for a particular task. More training data can be generated through synthetic modification of the available dataset. The advantage of doing this is that it makes the model robust to noise and invariant to certain properties e.g., translation, illumination, size, rotation etc. This is especially useful when working with satellite datasets as they usually contain few images. This is because training dataset is difficult or expensive to acquire as manual data labeling can be very time consuming.\

Training a model on a very small dataset may lead to poor performance when applied to a new, never-seen-before dataset. Augmentation techniques such as random flipping, rotation, zooming and channel averaging have been successfully applied to RGB images; however these techniques are limited when applied to satellite imagery and might not have a significant impact as a result of inherent uniformity in the images, as also reported by \cite{s21238083}. 
Despite the unique challenges associated with training deep learning models with images, it is widely accepted that data augmentation can lead to significant improvements in the performance of these models by making them more robust to noise and less likely to overfit.\

In addition, it is general knowledge in machine learning and deep learning that having more data has the potential of increasing model accuracy. Even in scenarios where there are lots of data available for training, data augmentation can still help to increase the amount of relevant data in the dataset. We therefore propose the use of a generative model using Generative Adversarial Networks (GANs) which can produce realistic satellite images to improve the performance of deep learning models on the task of Land Use and Land Cover classification. 
The rest of this paper is organized as follows. The related works and similar literature are presented in Section 2. Section 3 describes the dataset used in this work. The baseline model selected and information about the techniques used to achieve the baseline are described in Section 4. The model we are proposing is presented in Section 5 while the experiments performed using GANs are presented in Section 6. Sections 7 and 8 have the results and discussion respectively.  

\section{Related Works}

\subsection{Augmentation using Geometric Techniques}

Data augmentation increases the amount of data available to train a model. The primary advantage of doing this is that it makes the model more robust and less susceptible to overfitting\cite{Dave2020-zz}\cite{Ghaffar2019-nw}. In addition, it can improve the model performance by mimicking the image with different features \cite{Lei2019-er}. Different augmentation techniques have been shown to improve the model performance differently, depending on the quantity and quality of these features.\

Small satellite image datasets have been used for training deep learning models in existing works. For instance, \cite{Wicaksono2019-fo} and \cite{Saralioglu2022-oe} used a single image for classification and semantic segmentation tasks respectively. However, data augmentation has become more popular in recent times. Several relevant augmentation techniques have been compared in \cite{Abdelhack2020ACO}. Horizontal and vertical flipping had the highest accuracy out of the techniques considered for classifying satellite images \cite{Abdelhack2020ACO}. Image zooming or scale augmentation was used by \cite{Lei2019-er}. Here, the image is zoomed in or out, depending on a rate magnitude. Rotation augmentation is another relevant technique in which several copies of an  image are produced by rotating it through different angles. Furthermore, the authors in \cite{Yu2017DeepLI} used flip, translation and rotation in remote sensing scene classification. However, \cite{michannel2021} concluded that geometric transformations like rotation, zooming and translation have limited use for medium and low-resolution satellite data as they do not provide enough variability.\

Existing literature in the domain of remote sensing have used multi-temporal satellite data for both classification and semantic segmentation. \cite{Persson2018-gv} shows that combining images from several years from the same sensor on a single location improves model performance. With multi-temporal data, \cite{Skriver2012-nl} found the best date of observation based on available data. The authors in \cite{8903738} combined images from different dates by taking a weighted average of the spectral values. \cite{Tompson2015-dc} proposed channel dropout as a means of reducing overfitting of a CNN model trained on RGB images. The technique involved setting the pixel value of a channel with some predefined probability to zero. Color Jittering is another data augmentation technique that has been successfully applied to RGB images. Here, the pixel values in each channel are multiplied by some random number within a fixed range. \cite{Taylor2017-nl} used this technique to augment an RGB image dataset.\

Recently, \cite{michannel2021} showed that a Mix Channel approach, where a channel from the original image is replaced with the same channel of another image of the same location on a different date, outperformed state of the art models. Mix Channel approach showed better generalizability compared to generic data augmentation techniques like color jittering and geometric transformations. Experiments from \cite{michannel2021} also showed that channel drop out had promising results. Even though it did not outperform the mix channel approach, it outperformed the baseline model significantly.\

\subsection{Augmentation with GANs}

Generative Adversarial Networks (GANs) have also been used for data augmentation on traditional RGB images and Satellite images. This is an unsupervised, way of generating data \cite{gautam2020realistic}. The generative model usually comprises of the generator and discriminator, which work like gaming components. The former learns to generate realistic images and trick the discriminator which attempts to differentiate between the real and synthesized images. Examples of GANs, which have  been  applied to satellite images are DCGAN \cite{radford2015unsupervised}\cite{gautam2020realistic}, CycleGAN \cite{ren2020cycle}, SSSGAN\cite{rs13193984}. In \cite{gautam2020realistic}, the authors identified the challenge of the discriminator loss tending towards infinity while the generator loss immediately tends towards zero when training. High resolution images were generated using a progressive growing GAN. \cite{Abady2020-ba} applied GANs to satellite images for both image generation and image style transfer with visually promising results. However, no attempt was made to use the synthetically generated images for classification or segmentation tasks.\

Conditional GAN has also been applied to satellite images in \cite{kulkarni2020semantic}. They observed a better performance for fully supervised training and they were able to achieve this performance with just a slight increase in number of parameters. Marta GANs was also used in \cite{Lin2017MARTAGU}. When compared to DCGAN, the Marta GAN is capable of generating images at a higher resolution. \

\subsection{Land Use and Land Cover Classification (LULC) using EuroSAT data}

Some similar works have been done on improving the performance in image classification for LULC. In \cite{Xu2013RemoteSI}, principal component analysis was used to improve the task and to reduce the redundancy of the remote sensing images. This outperformed the maximum likelihood method. \cite{basu2015deepsat} proposed a classification framework that extracts features from a remote sensing input image, normalizes and feeds them to a Deep Belief Network for classification. In \cite{liu2016active}, an active learning framework based on a weighted
incremental dictionary learning was proposed. This was compared with other active learning algorithms and their method was observed to be more effective and efficient. Classification based on pixel on object-classification is investigated in \cite{doi:10.1080/22797254.2020.1790995}. In this project, we take the work a step further by investigating how GAN generated images improve the LULC task. Two GAN models are considered namely Deep Convolutional Generative Adversarial Network (DCGAN) and Wasserstein Generative Adversarial Network with Gradient Penalty (WGAN-GP). From our research, although DCGAN is not very scalable in terms of generating images from a lower resolution to a higher resolution, we found that it does perform well on 64x64 images as is also confirmed by \cite{Gautam2020-yd} and does latent space representation well. Since the original dataset contained 64x64 images and this work aims to add the generated images back into the training dataset to evaluate performance, the decision was made to use DCGAN, as it performs well for the image resolution we were working with. The use of ProGAN which is a GAN architecture based on a progressively growing training approach that increases images from lower resolutions to higher resolutions was also discussed, however implementing ProGAN would mean increasing the dimensions for all the images in the original dataset as well. Our choice of WGAN-GP was mainly to address possible instability in the training process and to reach faster convergence and this was apparent in the amount of time it took to generate the WGAN-GP images compared to how long the DCGAN images took for the same number of epochs.

\section{Dataset}

The dataset used for this task is the EuroSAT deep learning benchmark for LULC classification \cite{helber2019eurosat}.The EuroSAT dataset consists of images captured by the advanced resolutions cameras on the Sentinel-2 satellite (year of release 2015). The Sentinel-2 satellite is a remote land monitoring system that operates under the European Space Agency. The Sentinel-2 satellite provides images that have two different types of spectral resolutions, that is, the number of channels for each feature. The Sentinel-2 satellite could produce images of varying numbers of channels, for example, the Red-Green-Blue (e.g., the 3-channels RGB) and the multispectral (13 channels). The Sentinel-2 captures the global Earth’s land surface with a Multispectral imager (MSI) that detects and monitors the physical characteristics of an area by measuring the object reflected and emitted radiation at a distance every 5 days duration, and would provide the datasets for the next 20 - 30 years of time. \

The EuroSAT dataset is a novel dataset for remote sensing and capturing land use changes. It consists of high-resolution satellite images of rural and urban scenes.  The datasets are patch-based and provide macro-level details of the mapped area. The EuroSAT dataset consists of 10 labeled classes of LULC classification namely Forest, Annual Crop, Highway, Herbaceous Vegetation, Pasture, Residential, River, Industrial, Permanent Crop, and Sea/Lake  \cite{helber2019eurosat}. Its high-resolution satellite images, therefore, provide a clearer representation of objects. It contains 27000 labeled geo-referenced image data, each of 64x64 pixels, with a spatial resolution of 10m, of size 3.2 Terabytes (about 1.6 TB of compressed images) produced per day, with about 2000-3000 datasets per class  \cite{helber2019eurosat}. \
 
Past works on LULC problems have utilized the multispectral version of the Sentinel-2 EuroSAT dataset for training, which had stable learning and demonstrated good performances in differentiating different classes, although very slow in training. The RGB version is widely used because of its fast training and acceptable accuracy performance, while also known for its instability while learning \cite{isprs-archives-XLIII-B3-2021-369-2021}, and this was what was used in this project.  \cite{naushad2021deep} established benchmarking accuracy of 99.17\% on the RGB channel size. 

The dataset used for this project can be found online at \footnote{\url{https://www.kaggle.com/datasets/raoofnaushad/eurosat-sentinel2-dataset}} . The GAN images generated were renamed and added to this dataset folder, and this was used to perform an ablation of experiments. The images from the Kaggle link are in jpg format, while the generated images are in png format.The images were grouped into folders by class for the image generation, and for the model training, the label of the image was extracted from the filename, which contains the class to which the image belongs. 

\section{Baseline Selection}
Two models are used for this project, namely VGG16 and Wide Resnet50. To ensure early convergence, pretrained weights are used. VGG16 and Wide Resnet50 are proposed in \cite{naushad2021deep}, after experimenting with different architectures.

In \cite{naushad2021deep}, four experiments are carried out using the VGG16 and Wide Resnet50. LULC is performed on the EuroSAT dataset, with augmentation and without augmentation for both models. It was reported by the authors that the wide Resnet50 model with augmentation performed best. Early stopping was implemented both in the baseline model and our experiments. The difference in the number of epochs is a reflection of how quickly the model converges and stops improving. This was also confirmed when the baseline was re-established. The results are presented in table 1.  

\begin{table}[H]
\caption{Baseline implementation} 
\smallskip
\centering 
\begin{tabular}{c c c c c } 
\hline\hline 
Model & (b)Epochs & (b)Accuracy (\%) & (e)Epochs & (e)Accuracy (\%)\\ 

\hline 
VGG16 Accuracy without augmentation  & 18 & 98.14 & 20  & 98.12 \\

VGG16 Accuracy with augmentation  & 21 & 98.55 & 24 & 98.65 \\

Wide Resnet50 Accuracy without augmentation & 14 & 99.04 &24 & 98.81\\

Wide Resnet50 Accuracy with augmentation & 23 & 99.17 & 21 & 99.20 \\

\hline 
\end{tabular}
\\ b =  baseline , e = our experiment 
\label{table:nonlin} 
\end{table}

\section{Model Description}
\begin{figure}{H}
    \centering
    \includegraphics[width=14cm]{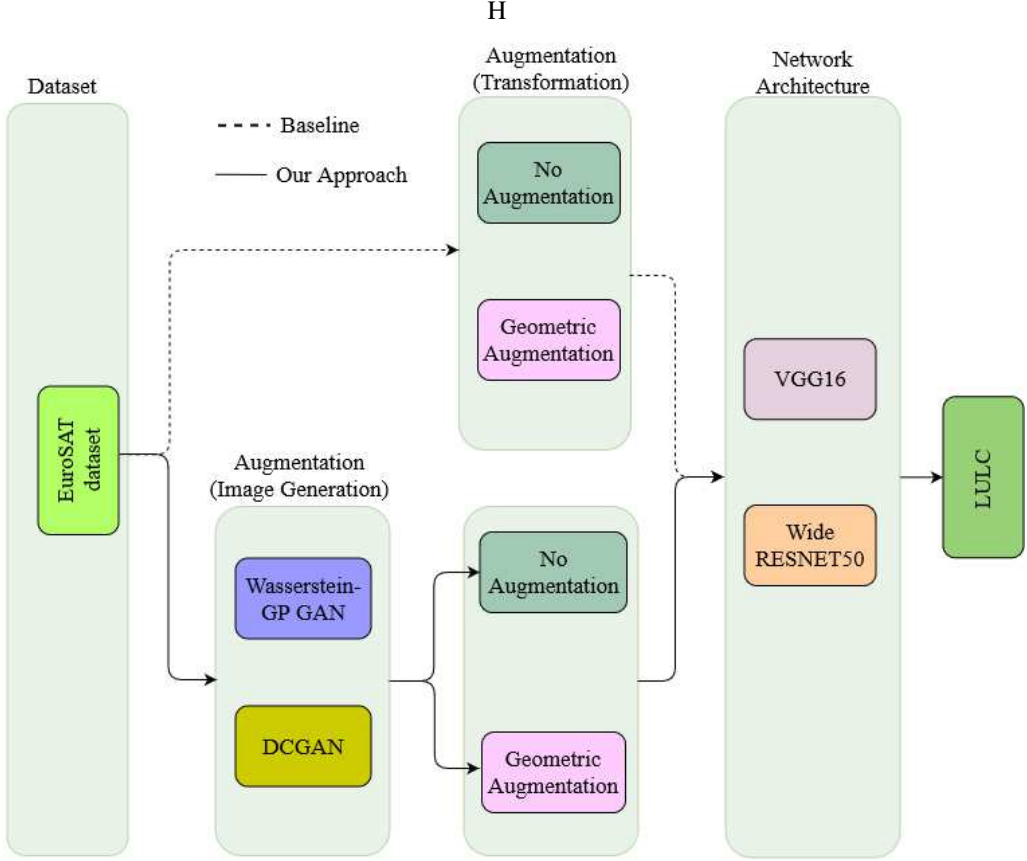}
    \caption{Work-layout of the project} 
\end{figure}

Figure 1 shows different experiments carried out, with and without data augmentation. The augmentation used for this project are geometric, DCGAN and Wassertein GP generated images. These augmentation techniques are used to get more training dataset so the model does not overfit. It was observed that the training accuracy reaches 100\% accuracy when training with the Wide Resnet50 model. However, the validation accuracy was 98.81 \%.

\subsection{Mathematical Model}
\subsubsection{GAN}
The generative Adversarial network is modelled mathematically as a two-player game between the Generator (G) and Discriminator (D), where given a training set, the generator attempts to generate new data with statistics similar to those in the training dataset while the discriminator tries not to be fooled by differentiating the fake data from the real ones. The min-max game equations are adapted from \cite{curto2017high} and entail the following objective function\

\begin{equation} 
\min\limits_{{(G)}}\max\limits_{{(D)}}V(D,G) = E_{x\sim P_{data}}[\log D(x)]+ E_{z\sim P_z} [\log(1-D(G(z)))]
\end{equation}

Where x  is a ground truth image sampled from the true distribution  p\_data and z is a noise sample sampled from p\_z (that is uniform or normal distribution ).  G and D are parametric function where G\:p\_z→p\_data maps sample from noise distribution  p\_z to data distribution p\_data.
The goal of the Discriminator is to minimize :

\begin{equation}
    L^{(D)} = -\frac{1}{2}E_{x\sim p_{data}}[\log D{(x)}]-
 - \frac{1}{2}E_{z\sim p_{z}}[\log (1-D(G(z))))]
\end{equation}

If we differentiate it  w.r.t  D(x) and set the derivative  equal to zero, we can obtain the optimal strategy  as: 
\begin{equation}
D(x)=  \frac{p_{data} (x)}{p_{z(x)}+ p_{data}(x)}
\end{equation}
This can be understood as follows,  the input is accepted  with its probability evaluated  under the distribution of the data,  $p_{data}(x)$ and then its probability is evaluated under the generator distribution of the data, $p_{z}$. Under the condition in D of enough capacity, it can achieve its optimum. It should be noted that the discriminator does not have access to the distribution of the data, but it is learned through training. The same  applies for the generator distribution of the data. Under the conditions in G of enough capacity, then it will set  $p_{z}= p_{data}$. The result is \begin{equation}D(x)=  \frac{1}{2}\end{equation} which  is actually the Nash equilibrium. Therefore, the generator  is referred to as the perfect generative model, sampling from p(x).

\subsubsection{DCGAN}
DCGAN mathemical model is similar to that of the normal GAN. However, DCGAN is different because it uses convolutional layers to facilitate stability during model training.  Also, to replace the fully connected layer, the generator upsamples using transposed convolution layers. Rectified linear unit (ReLU) is used in all the layers except the output layer which uses a hyperbolic tangent function (Tanh). 

Likewise, in the discriminator, strided convolutional layers are used.The discriminator downsamples using  convolutional layers with stride, instead of maxpooling. In contrast, the Leaky ReLU is used in all the layers. Furthermore, batch normalization is used for the generator and the discriminator.

The generator and discriminator loss functions are given by

\begin{equation}
     L_G^{(DCGAN)} = E_{z \sim{P_{z}(z)}}[\log D(G(z))]
\end{equation}

\begin{equation}
    L_D^{(DCGAN)} = E_{x\sim P_{data}(x)}[\log D_{(x)}]+ E_{z\sim P_{z}(z)}[\log(1- D(G(z))]
\end{equation}

\subsubsection{WGAN-GP}
Wasserstein GAN tackles the problem of GANs' loss functions being susceptible to hyperparameter choice and random initialization. It minimizes an approximation of the Wasserstein-1 distance which is the earth mover distance, between the distribution of the real images and the images generated by the GAN. The weights of the discriminator must be a K-Lipshitz function. This means that the first derivative of the function is bounded everywhere, and less than a constant.
Wasserstein GAN with gradient penalty (WGAN-GP) is an update to the firstly introduced Wasserstein GAN. WGAN-GP introduces gradient penalty approach, which solves the vanishing and exploding gradient problem of early WGAN. The vanishing and exploding gradient problem in early WGAN was as a result of weight clipping used. 

The generator and discriminator loss functions are:

\begin{equation}
     L_G^{(WGAN\_GP)} = - E_{z \sim P_{(z)}}[ D(G(z))]
\end{equation}

\begin{equation}
L_D^{(WGAN\_GP)} = E_{x\sim P_{data}(x),z\sim P_{z}(z)}[D(\widetilde{x})-D(x)+ \lambda {(\parallel \nabla x D({\tilde{x})} \parallel x-1)}^2)
\end{equation}

where $\epsilon \sim U[0,1],\widetilde{x} = G(z), \tilde{x} = \epsilon x + (1- \epsilon) \widetilde{x}$, 

The gradient penalty served as an efficient weight constraint to enhance the stability of the WGAN over the weight clipping \cite{gulrajani2017improved}. It is the second half of equation 8. It is a differentiable function that enforces the Lipschitz constraints by penalizing the gradient norm of the  critic’s output  for  random samples \cite{gulrajani2017improved}.  The properties of the optimal WGAN critic is also given in \cite{gulrajani2017improved}.  

\subsection{VGG16 and ResNet50}

\begin{figure}[H]
    \centering
    \includegraphics[width=14cm, height= 8cm]{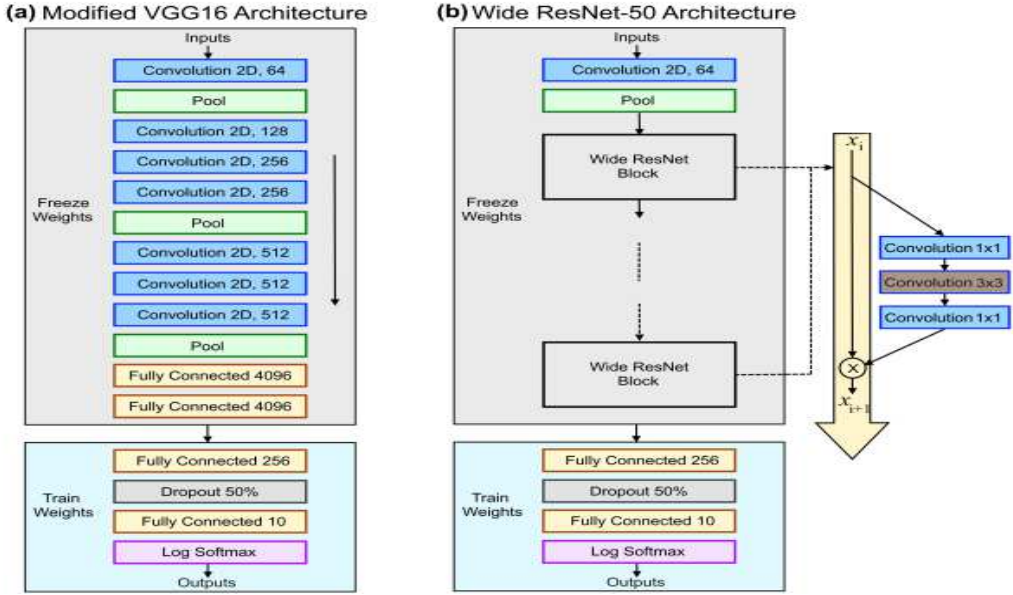}
    \caption{ Model architectures (a) Modified VGG16 architecture with training and freezing layers, and (b) Wide ResNet-50 architecture with training and freezing layers \cite{naushad2021deep}}
    
\end{figure}

VGG16 is a deep convolutional network proposed by K. Simonyan and A. Zisserman from the University of Oxford. The model achieved 92.7 \% top 5 test accuracy in ImageNet, with dataset of over 14 million images belonging to 1000 images.The convolutional block processes multiple convolutional layers. The top and bottom layers learn low level and high level features respectively. The ResNet architecture is an adaption of convolutional neural networks to tackle the problem of vanishing/exploding. In this technique, the model skips connections. The advantage of this is that any layer that makes the architecture to perform poorly is skipped. The ResNet architecture performs considerable well on image classification tasks. There are different variants of the ResNet and for this project, the Wide ResNet-50 is used. For this project, and in both cases, pretrained weights are used and the pretrained model is modified to add fully connected and dropout layers. Also, ReLU and log-softmax activation functions are added. Both models, normalised images are expected in minibatches of 3-channel RGB images of shape (3 X H X W) where H and W are expected to be 224. Also, some techniques are used to enhance the model's computation time and performance. These techniques are early stopping, data augmentation and adaptive learning rates.

Early stopping stops the training at an arbitrary number of epochs when the model's performance stops improving during training. The early stopping criteria used in this project is one  which terminates the training as  soon as  generalization exceeds  certain  threshold \cite{Prechelt1998-ts}. We had splitted the   EuroSAT dataset into the train dataset, validation set   of proportions 3 : 1  (75/25) and each  model   trained  such that the models  stop  the training at an arbitrary number of epochs when the model's performance stops improving during training. The threshold for number of consecutive epochs for which the model does not improve is set to 3. This means that if the training does not improve for 3 consecutive epochs, the training terminates. Moreover, we had used the early stopping criteria, which has been proven to give better performance with deep networks generally. Together, these validates  that the gradient descent with early stopping is provably robust to deep neural networks, adversarial networks inclusive, which validates the empirical robustness of deep networks as well as widely adopted heuristics to eliminate overfitting \cite{arxiv.2007.10099}.\\

In addition, in this project, we implemented the adaptive learning rate method to enhance model performance. This model efficiency technique is based on the momentum method that adds an adaptive property. The adaptive characteristics is one which implements as a step size change with the degree of the cost function. It will maintain the power of the momentum method and add adaptive properties to it to make a specific percentage of learning, and overall, constructs a step size according to the amount of the cost function. This causes it to be as close as possible to the minimum value of the cost function for convergence. In other words, the learning rate, is slowly reduced as it approaches convergence \cite{app11020850}. 
To reiterate, the different augmentations used are geometric augmentation, DCGAN, WGAN-GP.

\subsection{Geometric Augmentation}
The geometric augmentation methods used are random horizontal and vertical flip and random rotation. It was not stated by the authors of the baseline the intuition for choosing this augmentation methods as against other options such as colour jittering, random crop and so on. However, so as to make a fair comparison with the baseline, the geometric augmentation methods used are retained. 

\subsection{Evaluation Metric}
The task is a classification task and the metric used is accuracy. This is a widely used metric for classification task. This was the same metric used in the baseline which makes it possible to compare our experiments with the baseline. 

The model predicts what class the image belongs to. The accuracy is the number of accurately predicted images divided by the total number of predicted images.

\begin{equation}
Accuracy =  \frac{\tilde{y}}{y}
\end{equation}
where $\tilde{y}$ = accurate predictions, and y is total predictions

Also, for GAN models, Frechet Inception Distance is widely used to assess the quality of generated images; however, we have opted not to use it in this case as we think the effect the generated images have on the classification model performance is a better test of the images generated for this particular task.

\section{Experiments}

To investigate whether our GANs generated satellite images can improve the generalizability of a classification model for LULC task, we designed two image generation experiments using Deep Convolutional GAN (DCGAN) and Wasserstein GAN with gradient penalty (WGAN-GP). The DC GAN training was done on NVIDIA Tesla P100 GPU available on Kaggle. The WGAN-GP training was done on AWS G52X large instance.
Our codes are available on: \footnote{\url{https://github.com/Oarowolo11/11785-Project}}
\subsection{DC GAN Training}
The dataset consists of 27,000 satellite images in 10 classes. Each image is a 64x64 RGB image. All images in the dataset are already of the same size. We normalize the images such that pixel values are mapped between (1, -1) because normalized images have been shown to improve GAN training \cite{kurach2019large}. Following the DC GAN paper \cite{radford2015unsupervised}, we initialized the weights of convolutional layers to have a zero mean and a standard deviation of 0.02, while Batchnorm layers were initialized from a normal distribution with a mean of 1.0 and standard deviation of 0.02. The generator is a deep convolutional network with 5 blocks where each block consists of a transposed convolution, batch norm and an activation layer. A tanh activation is used in the final layer because the images have been normalized to be between 1 and -1. The generator receives an input noise vector of 100 dimensions to generate a 64 by 64 RGB image.
The discriminator is also a deep convolutional network with 5 blocks. It performs binary classification of an input image into a fake or real class. We used binary cross entropy as the loss function for both networks. The learning rate was 0.0002 and Adam optimizer was used for both networks. Beta coefficients of 0.5 and 0.999 were also used.
We trained the model to generate images in each class separately. Each class of image in the training data had between 2000 to 3000 images. We trained each GAN model for 300 epochs. 256 images were generated for each class making a total of 2560. The generated images constituted about 10 \% of the original training dataset.
We visually inspected the quality of the images generated and observed the effect of batch size used on the quality of images generated. Batch sizes greater than 16 produced images of worse quality on visual inspection. This may be because at the training, exposure of the discriminator to many images (batch number of images) may make it overpower the generator leading to poorer generated images.
A sample image generated by DC for each class is presented in the figures below:

\begin{figure}[ht]
    \centering
    \includegraphics[width=14cm]{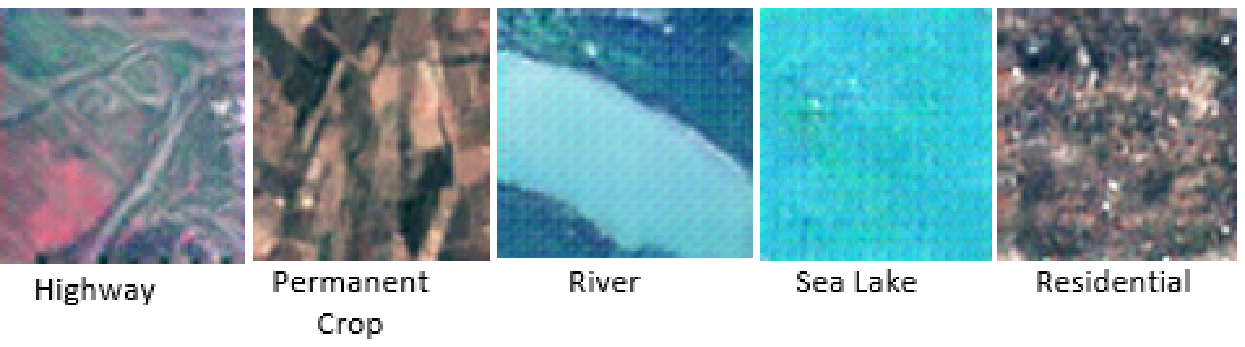}
    \includegraphics[width=14cm]{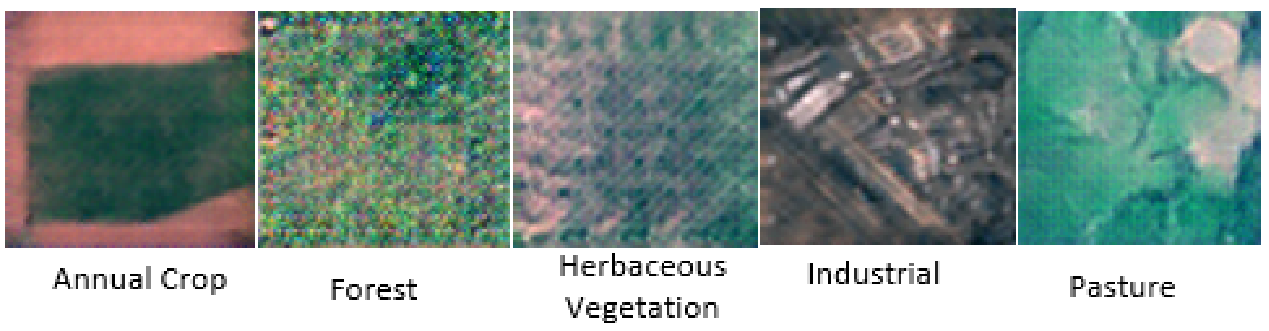}
    \caption{ Sample DCGAN Generated Images} 
\end{figure}
Tables 2 and 3 show the model structure for the generator and discriminator

\begin{table}[H]
\caption{DCGAN model structure-Generator} 
\smallskip
\centering 
\begin{tabular}{c c c} 
\hline\hline 
Layer(type)& Output Shape & Param \# \\ 
\hline 
ConvTranspose2d-1  & [-1,512,4,4] & 819,200 \\
BatchNorm2d-2& [-1,512,4,4] & 1,024 \\
ReLU-3   & [-1,512,4,4] & 0 \\
ConvTranspose2d-4 & 
[-1,256,8,8]& 2,097,152\\
BatchNorm2d-5& [-1,256,8,8] & 512\\
ReLU-6   & [-1,256,8,8] & 0 \\
ConvTranspose2d-7  & [-1,128,16,16] & 524,288 \\
\
BatchNorm2d-8& [-1,128,16,16] & 256\\
ReLU-9   & [-1,128,16,16] & 0 \\

ConvTranspose2d-10 & 
[-1,64,32,32]& 131,072\
\\
BatchNorm2d-11& [-1,64,32,32] & 128\\
ReLU-12   & [-1,64,32,32] & 0 \\

ConvTranspose2d-13 & 
[-1,3,64,64]& 3,072\\

Tanh-14 & 
[-1,3,64,64]& 0\\

\hline 
\end{tabular}

\label{table:nonlin} 
\end{table}

\begin{table}[H]

\caption{DCGAN model structure-Discriminator}  
\smallskip
\centering 
\begin{tabular}{c c c} 
\hline\hline 
Layer(type)& Output Shape & Param \# \\ 
\hline 
Conv2d-1  & [-1,64,32,32] & 3,072 \\
LeakyReLU-2   & [-1,64,32,32] & 0 \\
Conv2d-3 & 
[-1,128,16,16]& 131,072\\
BatchNorm2d-4& [-1,128,16,16] & 256 \\

LeakyReLU-5   & [-1,128,16,16] & 0 \\
Conv2d-6  & [-1,256,8,8] & 524,288 \\
\
BatchNorm2d-7& [-1,256,8,8] & 512\\
LeakyReLU-8   & [-1,256,8,8] & 0 \\

Conv2d-9 & 
[-1,512,4,4]& 2,097,152\
\\
BatchNorm2d-10& [-1,512,4,4] & 1,024\\
LeakyReLU-11   & [-1,512,4,4] & 0 \\

Conv2d-12 & 
[-1,1,1,1]& 8,192\\

Sigmoid-13 & 
[-1,1,1,1]& 0\\

Flatten-14 & 
[-1,1]& 0\\

\hline 
\end{tabular}

\label{table:nonlin} 
\end{table}

\subsection{WGAN-GP Training}
We trained a WGAN-GP to generate 256 images for each class. According to \cite{gulrajani2017improved}, this method does not necessarily generate better images than DCGAN approach, but it does provide advantage in form of better training stability. We used a similar generator architecture to the one used in DCGAN. We kept the learning rate and optimizer the same for comparison with DCGAN.
We used the code provided by \cite{radford2015unsupervised} to compare their results. The results obtained are presented in figure 4. Also, the architecture for the generator and critic are in table 4 and 5.

\begin{figure}[H]
    \centering
    \includegraphics[width=14cm]{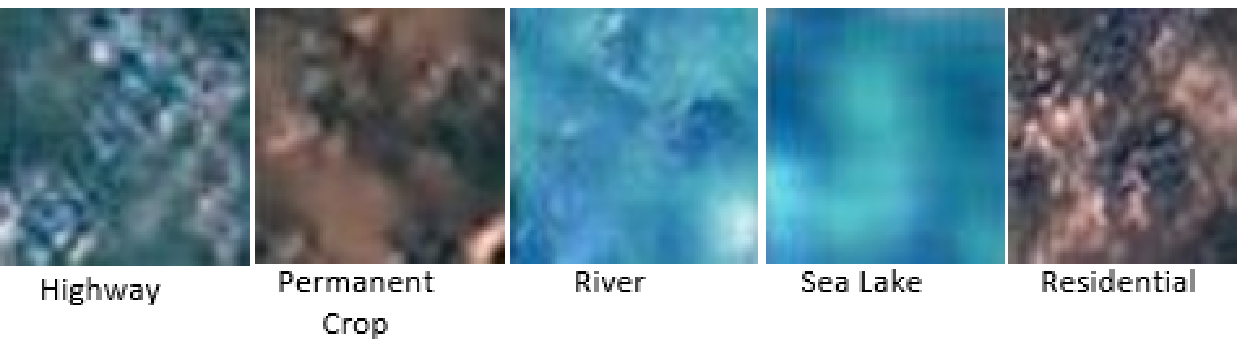}
    \includegraphics[width=14cm]{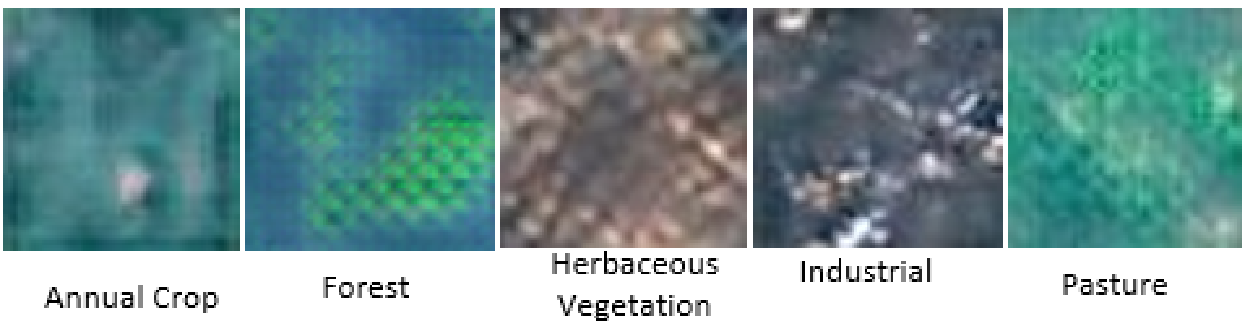}
    \caption{Sample of WGAN-GP Generated Images} 
\end{figure}

\begin{table}[H]
\caption{WGAN-GP model structure-Generator} 
\smallskip
\centering 
\begin{tabular}{c c c} 
\hline\hline 
Layer(type)& Output Shape & Param \# \\ 
\hline 
dense(Dense)  & (None, 32768) & 4227072 \\
relu (ReLu) & (None, 32768) & 0 \\
reshape (Reshape)   & (None, 8, 8, 512) & 0 \\
Conv2DTranspose & 
(None, 16,16,256) & 2,097,152\\
BatchNorm& (None, 16,16,256) & 1024\\
ReLU  & (None, 16,16,256) & 0 \\
ConvTranspose2d  & (None, 32, 32, 128) & 524,288 \\

BatchNorm & (None, 32, 32, 128) & 512\\
ReLU   & (None, 32, 32, 128) & 0 \\

ConvTranspose2d & 
(None, 64, 64, 4)& 8192
\\
BatchNorm2d & (None, 64, 64, 4) & 16\\
ReLU   & (None, 64, 64, 4) & 0 \\

ConvTranspose2d & 
(None, 64, 64, 3)& 195\\

\hline 
\end{tabular}

\label{table:nonlin} 
\end{table}

\begin{table}[H]
\caption{WGAN-GP model structure-Critic}  
\smallskip
\centering 
\begin{tabular}{c c c} 
\hline\hline 
Layer(type)& Output Shape & Param \# \\ 
\hline 
Conv2d  & (None, 32, 32, 64) & 3136 \\
leaky ReLu & (None, 32, 32,64) & 0 \\
Conv2D   & (None, 16, 16, 128) & 131200 \\
Leaky ReLU & 
(None, 16,16,128) & 0\\
Conv2D & (None, 8,8,128) & 262272\\
Leaky ReLU  & (None, 8,8,128) & 0 \\
Flatten  & (None, 8192) & 0 \\

dropout (Dropout) & (None, 8192) & 0\\
Dense   & (None, 1) & 8192 \\

\hline 
\end{tabular}

\label{table:nonlin} 
\end{table}

\section{Results}
\cite{naushad2021deep} trained four models using VGG16 and ResNET50 as pretrained base models. Four experiments were performed. First, each of the pretrained models was used for the classification without geometric augmentation, for the second round of experiments, random horizontal flip, random vertical flip, and random rotation were applied as geometric augmentations. We also repeated these experiments with the GANs generated images added to the original images. The results obtained are presented in table 6. It is observed that the different experiments train for different epochs because of the early stopping applied.

\begin{table}[h]
  \caption{Experiment results}
  \centering
  \begin{tabular}{lcccccc}
    \toprule
    \multirow{2}{*}[-0.5\dimexpr \aboverulesep + \belowrulesep + \cmidrulewidth]{Model}
    & \multicolumn{2}{c}{Baseline} & \multicolumn{2}{c}{DCGAN} & \multicolumn{2}{c}{WGAN-GP} \\
    \cmidrule(l){2-3} \cmidrule(l){4-5} \cmidrule(l){6-7}
    & Epochs & Accuracy & Epochs & Accuracy & Epochs & Accuracy \\
    \midrule
    VGG16 without augmentation & 18 & 98.14 & 14 & 98.17 & 15 & 98.2 \\
    VGG16 with augmentation & 21 & 98.55 & 25 & 98.52 & 25 & 98.38\\
    Resnet50 without augmentation & 14 & 99.04 & 21 & 98.81 & 18 & 98.88\\
    Resnet50 with augmentation & 23 & 99.17 & 21 & 99.15 & 25 & 99.12\\
    \bottomrule
  \end{tabular}
\end{table}

Some of the generated images are presented in figure 3 and figure 4.

\section{Discussion}
The results show that the GANs augmented dataset achieved comparable performance to the original dataset. The type of GANs architecture however seem to have no obvious effect on the model performance. This may be attributed to the fact that the number of images added to each dataset is just about 10\%. While there is not a significant difference in the model accuracy with GANs images, the important thing is that the model performance did not worsen, which shows that the GANs images in both cases are of comparable quality to the actual dataset. GANs images could however show significant improvement in model performance for a smaller dataset. Smaller datasets are more sensitive to the increase in their size, and a classification model trained with a smaller dataset may see the biggest positive impact from the use of GAN generated images.

Also, geometric augmentation has good effects on model accuracy in every experiment. A combination of geometric augmentation and GANs generated images could be useful where data is limited as GANs images could reduce the tendency of the model to overfit when geometric augmentation alone is used. 

Furthermore, the already high accuracy of the baseline models makes it especially difficult to improve these results. However, our work shows that augmenting the base dataset with GANs generated images does not worsen the performance of the classification models. Hence, our generated images are of sufficient quality for the task. This can improve generalizability of the classification models and will prove especially useful when limited training data is available.

\section{Conclusion}
GANs generated satellite images can improve the generalizability of deep classification models. We used two GAN architectures, DCGAN and WGAN-GP, to generate artificial satellite images. The type of architecture used had no apparent effect on model performance. The main advantage WGAN-GP offers from literature is training stability when training with deeper residual networks. In this case, it offers no advantage over DCGAN. Geometric augmentation can be used in combination with GAN augmentation for improved model performance. 

For future research, the effect of GANs generated images on a smaller dataset can be investigated. The effect of GANs augmentation for image datasets with severe class imbalance can also be explored. Finally, determining the effect of GANs generated images for other tasks such as semantic segmentation and built structure counting can also be considered. 

\section{Division of work}

Members of the group contributed to different aspects of the project, both technical and non-technical essentials. The search for the baseline was done by every member of the group. Also, every member contributed to report writing and editing. Specifically, Olayiwola Arowolo contributed to writing the literature review, experiments, results and discussion. He also actively participated in generation of images using DCGAN, and setting the environment up with dataset for image generation. Olayiwola also managed the Git repository for the project. Peter Owoade researched GANs and probable types to consider. He also researched the implementation of different GANs. He actively participated in generating images using WGAN-GP. He also contributed to reestablishing the benchmark model. He was also in charge of correspondence with the project mentor. Peter also made the presentation slides for the project review. Opeyemi Ajayi researched GANs and probable types, actively contributed to writing the Dataset, introduction and mathematical model sections of the project. She also generated some images using DCGAN and the restablishment of the baseline model. She setup the work environment which was used for training. She was also in charge of setting up meeting links, and reminders to keep the team on track. Oluwadara Adedeji was in charge of manuscript writing in latex. He also contributed to the literature review, baseline, and mathematical model section of the report. He participated in image generation using WGAN-GP and the restablishment of the baseline model. He contributed to designing the workflow and layout of images and tables in the report. He was also in charge of correspondence with the TA mentor.

\section{Acknowledgement}
We would like to thank Jacob Lee (project mentor) and Iffanice Houndayi (TA mentor) for providing valuable advice during this project.

\maketitle

\bibliographystyle{IEEEtran}
\bibliography{bibliography}

\end{document}